\documentclass[3p,times,procedia]{elsarticle}
\usepackage{arabtex}
\usepackage{utf8}
\setcode{utf8}
\usepackage{ecrc}
\usepackage[bookmarks=false]{hyperref}
    \hypersetup{colorlinks,
      linkcolor=blue,
      citecolor=blue,
      urlcolor=blue}

\volume{00}

\firstpage{1}

\journalname{Procedia Computer Science}

\runauth{Hammouda et al.}
\jid{procs}

\usepackage{pifont}

\usepackage[figuresright]{rotating}

\usepackage{times}
\usepackage{latexsym} 
\usepackage{microtype}
\usepackage{arabtex}
\usepackage{utf8}
\usepackage{xspace}
\usepackage{inconsolata}
\usepackage{algorithm}
\usepackage{algpseudocode}
\setcode{utf8}
\usepackage{xspace}
\usepackage{makecell}

\usepackage[colorinlistoftodos,prependcaption,textsize=tiny]{todonotes}

\newcommand{\TrAr}[1]{\arabtrue\transfalse {\scriptsize \Ar{#1}} /\arabfalse\transtrue \RL{#1}\arabtrue\transfalse}
\usepackage[T1]{fontenc}
\usepackage[utf8]{inputenc}
\usepackage{dingbat}
\usepackage{multirow}
\usepackage[colorinlistoftodos,prependcaption,textsize=tiny]{todonotes}
\usepackage{caption}
\newcommand{\tag}[1]{{\small#1}}
\newcommand{\Ar}[1]{{\scriptsize\<#1>}}
\newcommand{\sinatools}{\cal{\textit{SinaTools} }}
\newcommand{\alma}{\cal{\textit{Alma}}\xspace}
\newcommand{\salma}{\cal{\textit{Salma}}\xspace}
\newcommand{\qabas}{\cal{\textit{Qabas}}\xspace}

\begin{document}
\begin{frontmatter}

\dochead{6th International Conference on AI in Computational Linguistics}

\title{\sinatools: Open Source Toolkit for Arabic Natural Language Processing}
\author[a]{Tymaa Hammouda} 
\author[a]{Mustafa Jarrar\corref{cor1}}
\author[a]{Mohammed Khalilia}
\address[a]{Birzeit University, Birzeit, PO Box 14, West Bank, Palestine}

\begin{abstract}
We introduce \sinatools, an open-source Python package for Arabic natural language processing and understanding. \sinatools is a unified package allowing people to integrate it into their system workflow, offering solutions for various tasks such as flat and nested Named Entity Recognition (NER), fully-flagged Word Sense Disambiguation (WSD), Semantic Relatedness, Synonymy Extractions and Evaluation, Lemmatization, Part-of-speech Tagging, Root Tagging, and additional helper utilities such as corpus processing, text stripping methods, and diacritic-aware word matching.
This paper presents \sinatools and its benchmarking results, demonstrating that \sinatools outperforms all similar tools on the aforementioned tasks, such as Flat NER ($87.33\%$), Nested NER ($89.42\%$), WSD ($82.63\%$), Semantic Relatedness ($0.49$ Spearman rank), Lemmatization ($90.5\%$), POS tagging ($97.5\%$), among others. \sinatools can be downloaded from (\href{https://sina.birzeit.edu/sinatools}{https://sina.birzeit.edu/sinatools}).
\end{abstract}

\cortext[cor1]{Corresponding author. Tel.: +970-2-2982000}
\end{frontmatter}

\email{mjarrar@birzeit.edu}

\section{Introduction}
Despite the progress in Arabic NLP \citep{DH21}, there remain a lack of tools and resources that offer solutions for Arabic NLP and NLU tasks. Developing machine learning tools is crucial in democratizing Arabic NLP, as they allow people to incorporate machine learning into their workflows with less technical knowledge.
The availability of open-source and advanced AI tools remains limited. Low-code platforms and toolkits can bridge this gap by offering intuitive interfaces and trained models, making it easier for people in industry, research, and education to tap into the power of NLP to develop and deploy NLP applications. 

As will be discussed in this paper, a handful of tools for Arabic NLP have emerged, each offering rich functionalities that contribute to the growing ecosystem of Arabic NLP. Notable examples include The Stanford CoreNLP Toolkit \citep{manning2014stanford}, Farasa \citep{darwish2016farasa}, MADAMIRA \citep{pasha2014madamira}, CAMeLTools \citep{obeid2020camel}, and  OCTOPUS \citep{elmadany2023octopus}.

This article presents a new set of tools packaged in \sinatools, an open-source toolkit for Arabic NLP and NLU, offering state-of-the-art solutions for various semantic-related tasks developed by \href{https://sina.birzeit.edu/}{SinaLab} at Birzeit University. \sinatools currently supports flat, nested, and fine-grained Named Entity Recognition (NER), Word Sense Disambiguation (WSD), Semantic Relatedness, Synonymy Extraction and Evaluation, Lemmatization, Part-of-Speech (POS) tagging, and root tagging, among others. \sinatools provides a single end-to-end system for these tasks with different interfaces, including a Command Line Interface (CLI), Software Development Kit (SDK), and Application Programming Interface (API), as well as Python Jupyter Notebooks. Our goal is to simplify the development and deployment of Arabic NLP applications. Additionally, this article presents our benchmarking results of similar toolkits on the aforementioned tasks, demonstrating the superiority of \sinatools over other tools.  The main contributions of this article are:

\begin{enumerate}
    \item Open-source\footnote{Download page: \url{https://sina.birzeit.edu/sinatools}} and Python-based Arabic toolkit for various NLU tasks, with different interfaces.

    \item Benchmarking Arabic NLP toolkits, demonstrating \sinatools benefits and superior performance on all tasks. 
    
\end{enumerate}

This article is structured as follows: Section \ref{sec_related} overviews related work. Section \ref{sec_desgin} presents the design and implementation. Section \ref{sec:modules}
presents all NLU modules. Section \ref{sec_conclusion} concludes the paper and outlines our future work.

\begin{figure*}[t!]
    \centering
    \includegraphics[width=0.3\textwidth]{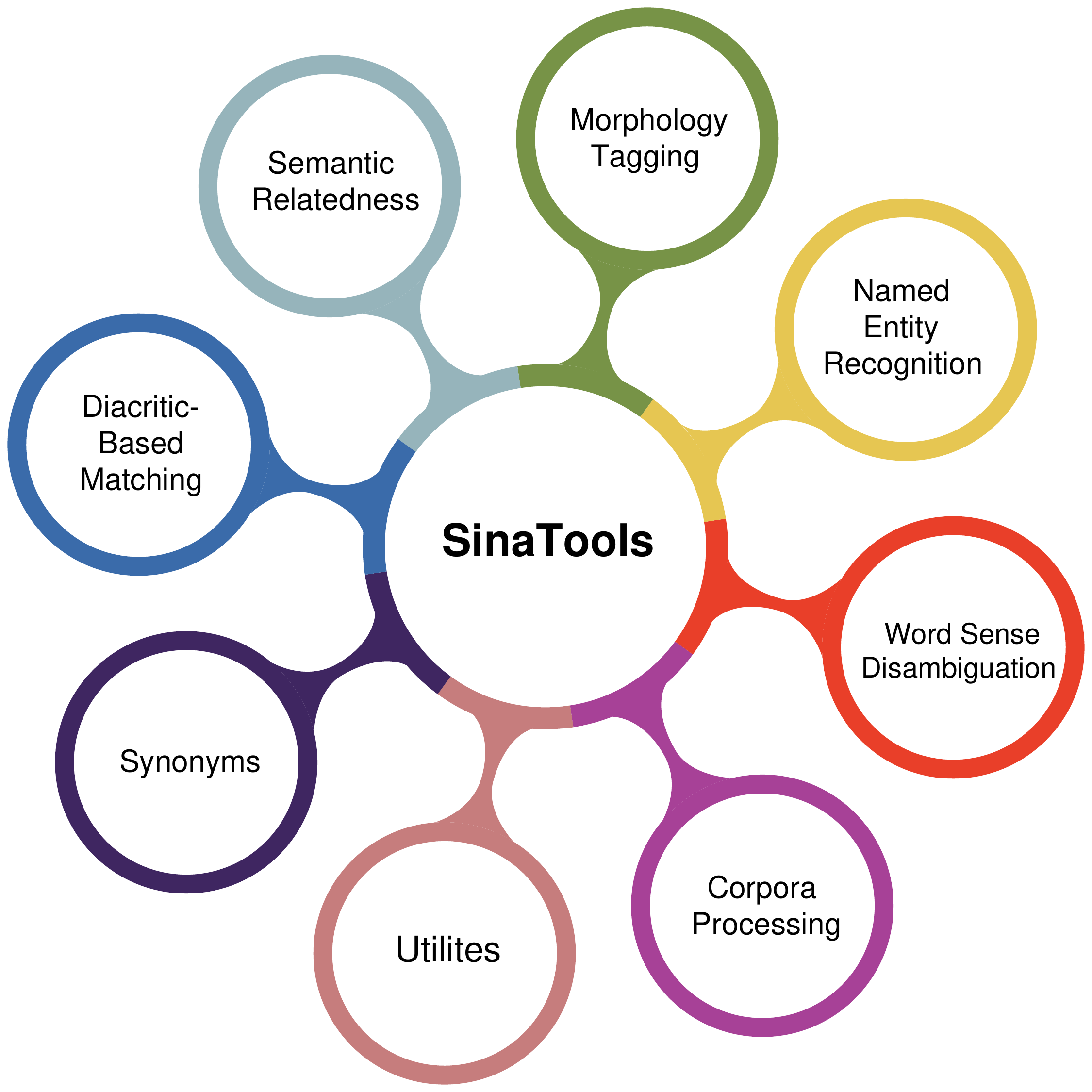}
    \caption{Core modules of \sinatools}
\label{fig:sinaToolsStructure}
\end{figure*}

\section{Related Work}
\label{sec_related}

This section overviews different Arabic NLP toolkits, which we also summarize in Table \ref{tab:compare_tools}. 
The Stanford CoreNLP  \citep{manning2014stanford} is a general-purpose toolkit, supporting nine languages including Arabic, but its Arabic support is limited to basic tasks.
MADAMIRA \citep{pasha2014madamira} is designed for morphological analysis and disambiguation, including lemmatization, POS tagging, and segmentation. As will be demonstrated later, MADAMIRA is among the top-performing tools in terms of accuracy. However, it is not open-source, is notably slow, and its functionalities are limited to morphology tasks.
Farasa \citep{darwish2016farasa} focuses also on Arabic morphology tasks, mainly lemmatization, POS tagging, and segmentation. Farasa is designed for speed, offering fast morphological analysis. However, its results are less sophisticated compared to MADAMIRA. For instance, Farasa's POS tagger is limited to 20 tags, whereas MADAMIRA supports 40 tags. Additionally, Farasa’s lemmatizer returns ambiguous lemmas as it removes all diacritics.

CAMeLTools  \citep{obeid2020camel} is an open-source NLP toolkit providing utilities for pre-processing, morphological modeling, dialect identification, NER, and sentiment analysis. Although CAMeLTools covers a broad range of tasks, its support of NLU tasks is limited. For example, its NER module can only detect three flat entity types: \tag{PERS}, \tag{LOC}, and \tag{ORG}.

OCTOPUS \citep{elmadany2023octopus} introduces a significant advancement with its AraT5v2 model, specifically designed for Arabic text generation. Meticulously trained on extensive and diverse datasets, the tools outperform competitive baselines across various tasks. OCTOPUS supports eight Arabic generation tasks, including summarization, paraphrasing, grammatical error correction, and question generation, among other generative tasks.

There are other task-specific tools. For example, TURJUMAN \citep{nagoudi2022turjuman} is a neural machine translation toolkit that translates from 20 dialects into MSA using the AraT5 model. AraNet \citep{abdul2020aranet} supports various social media tasks, including age, dialect, gender, emotion, irony, and sentiment prediction. Mazajak \citep{farha2019mazajak} focuses on sentiment analysis.

\begin{table*}[ht!]
\scriptsize
\centering
\captionsetup{justification=centering}
\caption{Feature comparison of related Arabic NLP tools.}
\begin{tabular}{|l|c|c|c|c|c|c|}
\hline
 &  \textbf{\sinatools} &  \textbf{CoreNLP} & \textbf{Farasa}   & \textbf{MADAMIRA} & \textbf{CamelTools} & \textbf{Octopus}\\ 
\hline
Language  & Python & Java& Java   & Java & Python& Python\\ \hline
Command Line Interface (CLI) & \checkmark & \checkmark &  \checkmark  & \checkmark & \checkmark & \\ \hline
Application Programming Interface (API)  & \checkmark & \checkmark &   \checkmark & \checkmark & \checkmark & \checkmark \\ \hline 
Morphological Modeling  &  & &    & \checkmark  & \checkmark & \\ \hline
Morphological Disambiguation   &  & \checkmark &  \checkmark  & \checkmark & \checkmark& \\ \hline
Diacritization  &  & &   \checkmark &   \checkmark& \checkmark & \checkmark\\ \hline
Tokenization/Segmentation/Stemming   &  & \checkmark &  \checkmark  & \checkmark & \checkmark & \\ \hline
Lemmatization  & \checkmark  & &  \checkmark   &  \checkmark  & \checkmark   &  \\ \hline
POS Tagging & \checkmark & \checkmark &  \checkmark  &  \checkmark & \checkmark & \\ \hline
Root Tagging  & \checkmark & & & & & \\ \hline
NER & 21 tags (flat, nested) & $4$ tags flat &   & & $3$ tags flat  & \\ \hline
WSD & \checkmark & &    &  & & \\ \hline
Semantic Relatedness  & \checkmark & &    &  & & \\ \hline

Sentiment Analysis  &  & &    &  & \checkmark & \\ \hline
Dialect ID   &  & &    &  & \checkmark & \\ \hline

Title Generation  &  & &    &  & & \checkmark \\ \hline
QA   &  & &    &  & & \checkmark \\ \hline
Question Generation  &  & &    &  & & \checkmark \\ \hline
transliteration    & \checkmark & &    &  & \checkmark & \checkmark \\ \hline
grammatical error correction    &  & &    &  & & \checkmark \\ \hline
Paraphrase    &  & &    &  & & \checkmark \\ \hline
Summarization &  & &    &  & & \checkmark \\  \hline
Synonyms Extraction & \checkmark & &    &  & & \\  \hline
\hline
\end{tabular}

\label{tab:compare_tools}
\end{table*}

\section{Design and Implementation}
\label{sec_desgin}
\subsection{Design}

\sinatools is an open-source toolkit consisting of a collection of Python application programming interfaces (APIs) and their corresponding command-line tools, which encapsulate these APIs. It adheres to these core design  principles:
\begin{enumerate}
\item \textbf{Modularity}: Each function is encapsulated in its own module, allowing for independent development, testing, and maintenance. This structure facilitates the seamless addition of new features without impacting existing ones. 

\item \textbf{Extensibility}: The architecture is designed to be extensible, enabling users to easily integrate additional tasks or replace existing components with custom implementations. 
\item \textbf{User-Friendly APIs}:  \sinatools provides intuitive and consistent APIs that abstract the complexity of underlying algorithms and data structures. This ensures that users, regardless of their expertise in NLP, can leverage the full capabilities of the tools with a minimal learning curve.
\item \textbf{Performance Optimization}: The implementation emphasizes efficient processing of large datasets. Large models and datasets required for various tasks are loaded the first time they are used or through a specific implemented download command. This approach minimizes load times and ensures efficient memory usage during operations.
\end{enumerate}

\subsection{Implementation}

\sinatools is implemented in Python 3.10.8 and can be installed via pip install. Python was selected due to its ease of use and its widespread adoption for NLP and Machine Learning. \sinatools is designed to be compatible with Python version 3.10 on Linux, macOS, and Windows. Currently, \sinatools provides both an API and a command-line interface for each component, in addition to demo pages. It is important to note that \sinatools is under continuous development, with ongoing additions of new features and updates to existing ones. This paper reports on the current components, but updates and new components will be continuously published.

\section{\sinatools Modules}
\label{sec:modules}
\subsection{Morphology Module}
\label{sec_lem}

This module is an implementation\footnote{Demo page (\alma): \url{https://sina.birzeit.edu/alma}} of the \alma (\Ar{ألمى}) morphology tagger presented in \citep{JAH24}, which consists of three sub-modules: (1) lemmatizer (2) POS tagger, and (3) root tagger. These sub-modules utilize a pre-computed memory, consisting of a dictionary containing many wordforms and their morphological solutions. This dictionary is implemented as a Python hashmap, simplifying the lemmatization, POS, and root tagging tasks into straightforward lookup operations.
Each entry in the dictionary is a key-value pair, where the key is a wordform and the value is its corresponding morphological solution. A morphological solution consists of a <$lemma$, $POS$, $root$>, and the frequency of this solution. 
The \alma dictionary is frequency-based. We have gathered a large collection of word forms and their morphological solutions from lexicographic resources developed at SinaLab. Using these, we calculated the frequency of each solution (see \citep{JAH24}). The primary resource for building \alma is the \qabas lexicographic data graph \citep{JH24}, which includes about $58k$ lemmas linked with lemmas in the Arabic Ontology \citep{J21,J11}, $110$ Arabic lexicons \citep{ J20,JA19,JAM19,ADJ19}, and other annotated corpora and resources \citep{ANMFTM23,JZHNW23,EJHZ22,JHRAZ17,JHAZ14,JDF11,J08}. The \alma dictionary retains the most frequent solution on the top (i.e., default) solution.
For example, the wordform (\TrAr{ذهب}) can be a noun and a verb. However, as the verb form is more common, \sinatools consistently tags it as a verb, regardless of the context. Although this method is simple and out-of-context, our evaluations show that it is more accurate and significantly faster than others.

\begin{table}[ht!]
\centering
\small
\caption{Benchmarking Arabic lemmatizers and POS taggers using the ATB and \salma datasets}
\label{tab:almma_eval}
\begin{tabular}{|l|rr|rr|r|r|rr|}
\hline
\multicolumn{1}{|c|}{\multirow{2}{*}{Tool}} & \multicolumn{2}{c|}{\begin{tabular}[c]{@{}c@{}}Lemmatization\\ exact match\\ (F1-Score)\end{tabular}} & \multicolumn{2}{c|}{\begin{tabular}[c]{@{}c@{}}Lemmatization\\ without Numbers\\ (F1-Score)\end{tabular}} & \multicolumn{1}{c|}{\begin{tabular}[c]{@{}c@{}}POS\\ 40 tags\\ (F1-Score)\end{tabular}} & \multicolumn{1}{c|}{\begin{tabular}[c]{@{}c@{}}POS\\ 3 tags\\ (F1-Score)\end{tabular}} & \multicolumn{2}{c|}{\begin{tabular}[c]{@{}c@{}}Speed\\ (seconds)\end{tabular}} \\ \cline{2-9} 
\multicolumn{1}{|c|}{}                      & \multicolumn{1}{c|}{ATB}                           & \multicolumn{1}{c|}{\salma}                       & \multicolumn{1}{c|}{ATB}                             & \multicolumn{1}{c|}{\salma}                         & \multicolumn{1}{c|}{ATB}                                                                & \multicolumn{1}{c|}{ATB}                                                               & \multicolumn{1}{c|}{ATB}                & \multicolumn{1}{c|}{\salma}           \\ \hline
MADAMIRA                                    & \multicolumn{1}{r|}{84.2\%}                        & 85.5\%                                           & \multicolumn{1}{r|}{88.3\%}                          & 86.6\%                                             & 82.3\%                                                                                  & 97.5\%                                                                                 & \multicolumn{1}{r|}{1661.3}             & 388.57                               \\ \hline
CamelTools                                  & \multicolumn{1}{r|}{-}                             & -                                                & \multicolumn{1}{r|}{80.9\%}                          & 81.7\%                                             & 82.7\%                                                                                  & 98.6\%                                                                                 & \multicolumn{1}{r|}{13080.2}            & 646.06                               \\ \hline
Farasa                                      & \multicolumn{1}{r|}{-}                             & -                                                & \multicolumn{1}{r|}{-}                               & -                                                  & 62.4\%                                                                                  & 94.3\%                                                                                 & \multicolumn{1}{r|}{33.22}              & 15.07                                \\ \hline
\sinatools                                   & \multicolumn{1}{r|}{87.8\%}                        & 90.5\%                                           & \multicolumn{1}{r|}{91.5\%}                          & 91.2\%                                             & 92.7\%                                                                                  & 97.5\%                                                                                 & \multicolumn{1}{r|}{11.34}              & 1.19                                 \\ \hline
\end{tabular}
\end{table}

\textbf{Evaluation}: Table \ref{tab:almma_eval} presents the benchmarking results of lemmatization and POS tagging for four tools. We used two datasets: (1) the \textbf{LDC’s Arabic TreeBank} (ATB) \cite{MaamouriAtb2010}, which includes $339,710$ tokens with their morphological annotations, and (2) the \textbf{\salma dataset} \cite{JMHK23}, a more recent corpus containing $34,253$ tokens with their morphological and semantic annotations.
For lemmatization, we compared the results of all tools in two scenarios: ($i$) with the exact spelling of the lemmas, and ($ii$) after removing numbers from lemmas. Farasa's results did not match in either case, as it provides ambiguous undiacritized lemmas.
For POS tagging, we evaluated the tools using the ($i$) full set of 40 tags and ($ii$) a set of 18 tags (see details in \cite{JAH24}). For speed benchmarking, we conducted four runs for each tool on the same machine (24 CPU, 47G Memory, CentOS, 1.3T size) under identical experimental conditions. Excluding the first run, we averaged the speeds of the remaining three runs. \sinatools outperformed the other systems in both speed and accuracy across the two corpora. More detailed experimentation and comparisons of the four tools are reported in \cite{JAH24}.

\textbf{Out-of-Vocabulary}: 
As \sinatools relies on a dictionary, it cannot provide solutions for wordforms not included in its dictionary. However, our benchmarking showed that out-of-vocabulary (OOV) instances are not a significant issue. To address OOV cases, \sinatools integrates a fine-tuned BERT model for POS tagging, ensuring robust performance even when encountering words not present in the dictionary (See \cite{JAH24}).

\subsection{Named Entity Recognition Module}
\label{sec_ner}

\begin{figure}[ht!]
\centering
\includegraphics[width=0.7\textwidth]{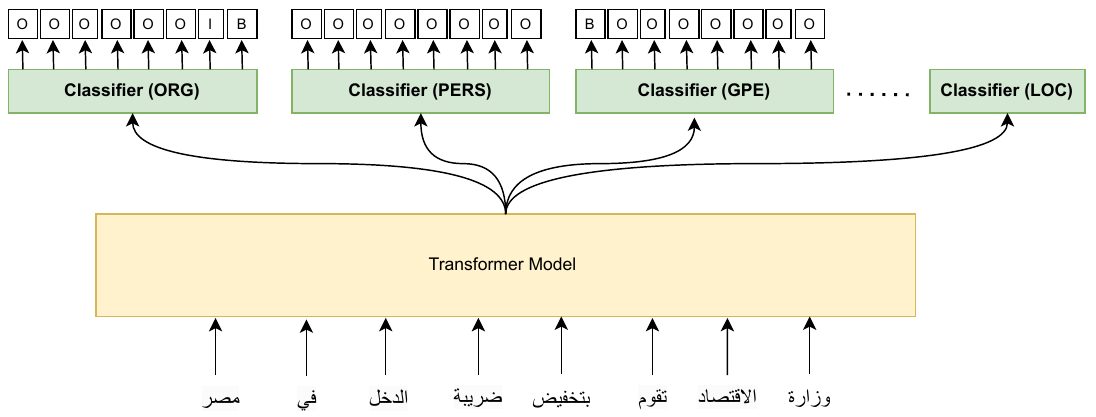}
    \caption{\sinatools NER model}
\label{fig:ner}
\end{figure}

This module\footnote{Demo page (Wojood): \url{https://sina.birzeit.edu/wojood}} is based on a BERT model that fine-tuned with our Wojood datasets \cite{JKG22}.
The module supports flat, nested, and fine-grain entity types. Since Wojood has 21 entity types, our model includes 21 classification layers, one layer for each entity type. Each layer classifies the token into one of three classes, $C=\{I, O, B\}$. As illustrated in Figure \ref{fig:ner}, each classifier is an expert in one entity type, which will output one of the three labels in $C$ for each token. 

\vspace{-0.25cm}

\textbf{Evaluation}: We benchmarked and compared our NER module with CamelTools on three datasets. 
Table \ref{tab:compare_ner} summarizes the results. \textit{Wojood} test set, which covers MSA, Palestinian, and Lebanese dialects, on which \sinatools is trained \cite{JKG22}. \textit{Wojood\textsuperscript{Gaza}} provided in subtask-3 in the WojoodNER 2024 shared task \cite{JHKTEA24}. This dataset contains $50$k tokens recently collected from five news domains (politics, law, economy, finance, health) related to the Israeli War on Gaza, and annotated using the $21$ Wojood tags. \textit{Politics}, a second out-of-domain dataset and contains $12,712$ tokens that we collected from Aljazeera news articles two years ago. We note that CamelTools can detect three types of entities only (\tag{PERS}, \tag{LOC}, and \tag{ORG}).  
\vspace{-0.25cm}

\begin{table}[ht!]
\centering
\caption{Evaluation of \sinatools NER module (F1-Score)}
\begin{tabular}{|l|l|r|c|c|} \hline
Dataset & Tool & Flat (3 tags) & Flat (21 tags) & Nested (21 tags) \\ \hline
\multirow{2}{*}{\textit{Wojood}} & CamelTools & 45.85\%     & - & - \\ 
& \sinatools & \textbf{66.40\%}    & \textbf{87.33\%} & \textbf{89.42\%} \\ \hline

\multirow{2}{*}{\textit{Wojood\textsuperscript{Gaza}}} & CamelTools & 29.72\%     & - & - \\ 
& \sinatools & \textbf{67.27\%}    & \textbf{55.72\%} & \textbf{62.68\%} \\ \hline

\multirow{2}{*}{\textit{Politics}} & CamelTools & 54.00\%     & - & - \\ 
& \sinatools & \textbf{69.00\%}    & \textbf{68.00\%} & \textbf{74.00\%} \\ \hline

\end{tabular}
\label{tab:compare_ner}
\end{table}

\subsection{Word Sense Disambiguation Module}
\label{WSD_section}
\begin{figure}[ht!]
    \centering
    \includegraphics[width=0.9\textwidth]{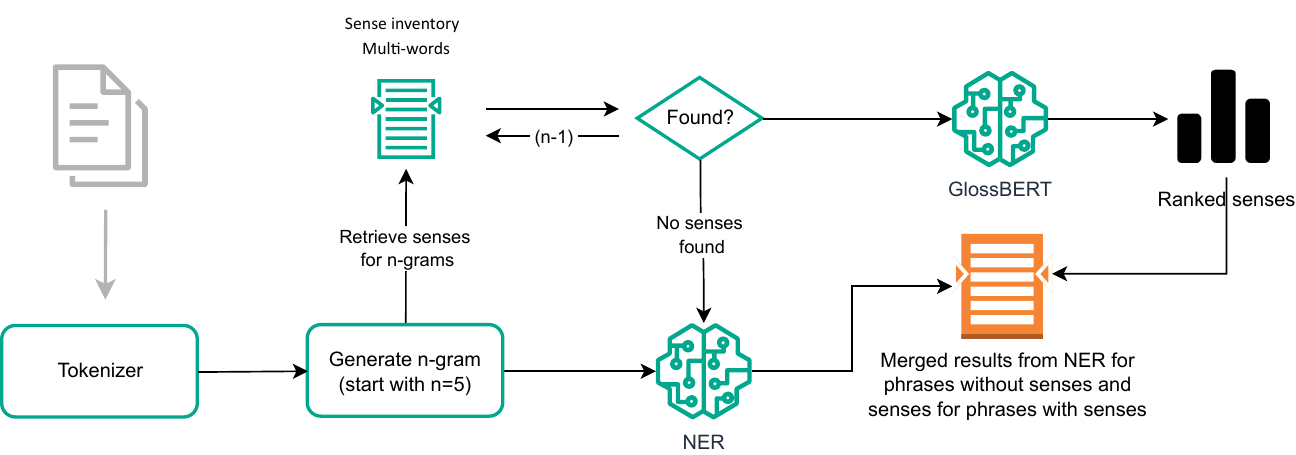}
    \caption{\sinatools end-to-end WSD (\salma)}
\label{fig:wsd}
\end{figure}

This module is an implementation of a novel end-to-end semantic analyzer called \salma (\Ar{سلمى})\footnote{Demo page (\salma): \url{https://sina.birzeit.edu/salma}}. Figure \ref{fig:wsd} illustrates our system architecture, as a pipeline of components: Tokenizer, Lemmatizer, POS Tagger, NER, Target Sense Verification (TSV), and two sense inventories. Given an input sentence, the WSD module conducts semantic analysis, which includes disambiguating single-word and multi-word expressions, and tagging named entities.

\textbf{Example}: Input (\Ar{وزارة الاقتصاد تقوم بتخفيض ضريبة الدخل في مصر} / \textit{Ministry of Economy is reducing income tax in Egypt}). Output: (1) \textbf{named entities} 
(\Ar{وزارة الاقتصاد}/Ministry of Economy)\textsubscript{ORG}
and (\Ar{مصر}/Egypt)\textsubscript{GPE}; 
(2) \textbf{multi-word expressions senses }(\Ar{ضريبة الدخل}/income tax)\textsubscript{\href{https://ontology.birzeit.edu/lexicalconcept/349001534}{\Ar{ضَرِيبَةٌ دَخْلٌ
}}}; and (3) \textbf{single-word senses}  (\Ar{تقوم}/is)\textsubscript{\href{https://ontology.birzeit.edu/lexicalconcept/303046074}{6-\Ar{قَامَ}}}, (\Ar{بتخفيض}/reducing)\textsubscript{\href{https://ontology.birzeit.edu/lexicalconcept/303015225}{2-\Ar{تَخْفِيضٌ}}}.

The WSD system consists of a pipeline of the following sub-processes:

\noindent\textbf{Phase 1 ($n$-gram tokenization)}:  we first generate $n$-grams from the input text for all $2 \leq n \leq 5$. The $n$-grams will be used in later phases to query the sense inventory. 

\noindent\textbf{Phase 2 (Lemmatization)}: Each token in a given $n$-gram is lemmatized using the \sinatools Lemmatizer.

\noindent\textbf{Phase 3 (Multi-word WSD)}:
At this phase we retrieve senses for multi-word expressions such as (\Ar{ضريبة دخل}/income tax) and (\Ar{سنة كبيسة}/leap year). \sinatools includes a dictionary of such expressions that we collected from our 150 lexicons \citep{JA19}, storing each multi-word expression with its glosses. This dictionary serves as our multi-word sense inventory.
For each lemmatized multi-word expression (i.e., $n$-gram, where $2 \leq n \leq 5$), we perform a lookup in the sense inventory, starting with $n=5$. If a sense is not found, we reduce $n$ by $1$ and attempt the retrieval again.
For example, (\Ar{ضريبة دخل}/income tax)  has two glosses (i.e., senses) in the sense inventory. The senses along with the original input sentence are sent to the TSV module for disambiguation (Phase 6). 

\noindent\textbf{Phase 4 (NER)}: As no need to sense-disambiguate all $n$-grams (e.g., text spans such as (\Ar{مصر}/Egypt) and (\Ar{وزارة الاقتصاد}/Ministry of Economy) are proper nouns), the NER module tags each span with entity types.

\noindent\textbf{Phase 5 (single-word WSD)}:
All tokens that were not sense-disambiguated or NER-tagged in the previous steps will undergo an additional step. We disambiguate them as single words ($uni$-grams). For instance, in the example above, three $uni$-grams need to be single-word disambiguated:  (\Ar{تقوم}/is), (\Ar{بتخفيض}/reducing), and (\Ar{في}/in). The glosses of these $uni$-grams are retrieved from the single-word sense inventory and sent to the TSV module for disambiguation (Phase 6).

\noindent\textbf{Phase 6 (Target Sense Verification (TSV))}:
We implemented the TSV approach proposed in \cite{JMHK23}, which is a core step in the WSD pipeline. 
Given a pair of sentences (context and gloss), TSV calculates the Positive and Negative probabilities to classify whether this pair is True or False, indicating whether the gloss is the correct sense used in the context based on the higher probability.
Thus, for a word $w$ in context $C$, and all $w$ glosses $g_i\in G$, we generate a set of pairs \{($C$, $g_1$),($C$, $g_2$),..., ($C$, $g_m$)\}. These pairs are sent to the TSV model for evaluation. The pair with the highest Positive probability is considered the target sense.  We fine-tuned the TSV model using AraBERT V2 \citep{antoun2020arabert} and the ArabGlossBERT dataset ($167k$ pairs) \citep{HJ21b}. We experimented with an augmented dataset \citep{MJK23} but it did not enhance the results. We also addressed WSD using the WiC model \citep{HJ21} but the performance was bad.

\textbf{Evaluation}: 
As \sinatools is the first tool for Arabic WSD, we computed its baseline using the \salma corpus \cite{JMHK23}, the only available Arabic sense-annotated corpus. \salma is only annotated with NER and single-word expressions, but we extend its annotations to include multi-word expressions. Table \ref{tab:WSD_baseline} shows that \sinatools achieves an overall WSD performance of 82.63\% accuracy. 
It's important to note that calculating the accuracy of the single-word WSD is not straightforward because a word might have multiple correct senses in \salma, and whether the accuracy should include words with single senses. Thus, we refer the reader to \cite{JMHK23} for more details. It is also worth noting that \salma was used in the ArabicNLU Shared Task \citep{KMSJAEZ24}, where the best-performing system on single-word WSD achieved 77.8\% \citep{rajpoot2024upaya}.

\begin{table}[h!]
\centering
\small
\caption{Evaluation of \sinatools WSD module} 
\begin{tabular}{|l|r|r|r|}
\hline
 & \textbf{Tokens Count}  &  \textbf{Span Count} 
 & \textbf{Accuracy}    \\ 
\hline

NER (6 types) & 4,389 &  2,728 & 85.31\%   \\ 
Multi-word WSD & 2,100  & 519 &  88.92\%  \\ 
Single-word WSD & $27,764$  & $27,764$ & $81.73$\% \\ 
\hline
\textbf{Overall} (Micro average) &   \textbf{$34,253$ }&  & \textbf{82.63\%} \\ 
\hline
\end{tabular} 
\label{tab:WSD_baseline}
\end{table}

\subsection{Semantic Relatedness Module}
\label{Semantic_section}

This task is useful in NLP for many scenarios such as evaluating sentence representations \citep{AsaadiMK19}, document summarization, and question answering. Given two sentences, the semantic relatedness task aims to assess the degree of association between two sentences across various dimensions, including meaning, underlying concepts, domain-specificity, topic overlap, or viewpoint alignment \citep{AbdallaVM23}. \sinatools supports MSA semantic relatedness, which represents our participation \citep{MJK24} in the SemRel Shared Task \citep{ousidhoum-etal-2024-semeval}, where we achieved the top rank. 
Unlike the lexical overlap (using dice coefficient) proposed in \citep{ousidhoum2024semrel2024}, we extracted the mean-pooling embeddings of the sentences from BERT-based model, then employed cosine similarity as an unsupervised technique to calculate the sentence-pair scores. We evaluated our method using the 595 sentence-pairs test set provided in the SemEval-2024 Task 1 on Semantic Textual Relatedness for African and Asian Languages \citep{ousidhoum-etal-2024-semeval}. Table \ref{tab:compare_str} shows that we outperformed the baseline. We used Spearman rank correlation coefficient, the official evaluation metric used in the shared task, which captures the level to which the system predictions align with the human judgments of the test pairs. 

\begin{table}[ht!]
\centering
\caption{Performance of Semantic Relatedness of \sinatools on the SemEval-2024 Task}
\begin{tabular}{|l|ll|}
\hline
\multirow{2}{*}{\begin{tabular}[c]{@{}l@{}}Test \\ Pairs\end{tabular}} & \multicolumn{2}{c|}{Spearman}  \\ \cline{2-3} 
  & \multicolumn{1}{l|}{Baseline} & \sinatools \\ \hline
595    & \multicolumn{1}{l|}{0.42}     & \textbf{0.49}$\uparrow$ \\ \hline
\end{tabular}
\label{tab:compare_str}
\end{table}
\vspace{-.25cm}

\subsection{Synonyms Module}
\label{Synonyms_section}
Arabic is low-resourced in terms of synonymy resources and tools \cite{KAJ21,JKKS21,APJ16}. \sinatools includes an implementation of the synonymy extraction and evaluation\footnote{Demo Page (Synonymy) : \url{https://sina.birzeit.edu/synonyms}} algorithm introduced in \cite{GJJB23}, which was also tested in extracting Welsh synonyms \cite{KAEMKRTM23}. The algorithm is designed to extract synonyms from mono and multilingual lexicons. It leverages synonymy and translation pairs from these resources to generate a synonymy graph, where nodes involved in cyclic paths are deemed synonyms. The extracted synonyms are assigned fuzzy values to indicate their degree of belonging to the synonym set.
The algorithm is evaluated using the Arabic WordNet, achieving 98.7\% accuracy at 3\textsuperscript{rd} level and $92\%$ at 4\textsuperscript{th} level. We utilized about $100$ mono and bilingual lexicons \cite{JA19}, including the Arabic Ontology \citep{J21,J11}, Qabas \citep{JH24}, $40$ ALECSO lexicons, $11$ from the Arabic Academy, WordNet, and others. These resources were used to compute two synonymy graphs: a 2\textsuperscript{nd} level graph ($75 MB$) and a  3\textsuperscript{rd} level graph ($1.1 GB$).
\sinatools features two main methods: (i) \textit{\textbf{SynExtract}} for synonym extraction, and (ii)\textit{ \textbf{SynEval}} for synonym evaluation. The $SynExtract$ method allows users to give one or more terms as input and retrieve their synonyms (See example 1). Each synonym is given a fuzzy value. The more terms are provided in the input the better the accuracy. The $SynEval$ method enables users to input a set of terms and receive a fuzzy score for each term, reflecting its synonymy strength (See example 2). In both methods, users can choose between the faster 2\textsuperscript{nd} level graph or the richer but slower 3\textsuperscript{rd} level graph.

\textbf{Example 1:} the function $SynExtract$(\Ar{طريق} , \Ar{ممر}), returns \{\Ar{مسلك} \textsubscript{61\%}, \Ar{سبيل} \textsubscript{50\%}, \Ar{نهج} \textsubscript{23\%}, \Ar{زقاق} \textsubscript{20\%}, \Ar{سكة} \textsubscript{15\%}\ , ...\}

\textbf{Example 2:} the function $SynEval$(\Ar{طريق} , \Ar{ممر}, \Ar{مسلك}), returns \{\Ar{مسلك} \textsubscript{70\%}, \Ar{ممر} \textsubscript{50\%}, \Ar{طريق} \textsubscript{30\%}\}.

\subsection{Other Modules and Tools}
\label{Others_section}

In addition to its core functionalities, \sinatools includes a variety of utility modules designed for text processing capabilities. The \textbf{Sentence Splitter} module offers the capability to segment text into sentences, with the flexibility to specify separators (e.g., periods, question marks, exclamation marks, and line breaks), thereby accommodating varied text structures. Notably, this feature selectively incorporates chosen separators while disregarding unselected ones, thereby enhancing the precision of text segmentation. The \textbf{Diacritic-Based Matching of Arabic Words} module\footnote{Demo Page (Diacritic-Based Matching) : \url{https://sina.birzeit.edu/resources/Implication.html}} compares two Arabic words to find out whether they are the same, taking into account their partial or full diacratizition \cite{JZAA18}.  For example, the two words (\TrAr{فَعلَ},\TrAr{فعَل}) are compatible and the (\TrAr{فَعلَ},\TrAr{فِعلَ}) are not. The \textbf{Text Duplicate Removal} module employs cosine similarity to eliminate redundant sentences from input text under a \textit{similarity threshold}, which is useful for corpora pre-processing. The  \textbf{Arabic Jaccard} module computes union, intersection, and similarity metrics between sets of Arabic words, taking into account partial diacritization. Moreover, the \textbf{Arabic Diacritic Removal (arStrip)} module designed to cleanse Arabic words by selectively removing diacritics, shaddah, digits, alif, and special characters, according to user-specified parameters. Lastly, the \textbf{Transliteration Module} facilitates seamless conversion between Arabic and Buckwalter transliteration schemas, thereby ensuring interoperability across different linguistic representations.

\section{Conclusions and Future Work}
\label{sec_conclusion}
We presented \sinatools, an open-source Python package for Arabic NLP, offering solutions for
various tasks such as NER, WSD, semantic relatedness, synonymy extraction and evaluation, lemmatization, POS
tagging, among others. Our benchmarking results highlight that \sinatools consistently outperforms similar tools across all tasks.

Looking ahead, we plan to expand \sinatools by incorporating additional Arabic NLU modules. These enhancements will focus on areas currently underserved by existing tools, such as intent detection \cite{JBKEG23,AraFinNLP24}, relationship extraction \citep{JDJK24}, detection of hate speech \cite{HJKN23}, bias and propaganda \cite{DJJQQ24,ZJHBZDEA24}, among others.

\bibliography{MyReferences_TEMP,bibliography}
\bibliographystyle{elsarticle-harv}

\clearpage
\normalMode
\end{document}